%% file: main.tex
\documentclass{article}
\usepackage{spconf,amsmath,graphicx}

\usepackage{caption}
\usepackage{subfigure}
\usepackage{amsfonts}
\usepackage{multirow}
\usepackage{mathtools}
\usepackage{xcolor}
\usepackage{comment}
\usepackage[normalem]{ulem}

\usepackage{algorithm}
\usepackage[noend]{algpseudocode}

\def\L{{\cal L}}

\title{Video action recognition via neural architecture searching}
\name{Wei Peng$^1$ \qquad Xiaopeng Hong$^{2,1}$ \qquad Guoying Zhao$^{1,\star}$\thanks{$^{\star}$ Corresponding author.~~This work has been accepted by 2019 26th  IEEE International Conference on Image Processing (ICIP 2019).}\thanks{\textcolor{blue}{Copyright @ 2019 IEEE. Personal use of this material is
permitted. However, permission to use this material for any other purposes must be obtained from the IEEE.}}}

\address{ $^1$Center for Machine Vision and Signal Analysis, University of Oulu, Finland\\
     $^2$ Xi'an Jiaotong University, Xi'an, P. R. China}

\begin{document}
\input{macros.tex}

\maketitle

\begin{abstract}
 Deep neural networks have achieved great success for video analysis and understanding. However, designing a high-performance neural architecture requires substantial efforts and expertise. In this paper, we make the first attempt to let algorithm automatically design neural networks for video action recognition tasks. Specifically, a spatio-temporal network is developed in a differentiable space modeled by a directed acyclic graph, thus a gradient-based strategy can be performed to search an optimal architecture. Nonetheless, it is computationally expensive, since the computational burden to evaluate each architecture candidate is still heavy. To alleviate this issue, we, for the video input, introduce a temporal segment approach to reduce the computational cost without losing global video information. For the architecture, we explore in an efficient search space by introducing pseudo 3D operators. Experiments show that, our architecture outperforms popular neural architectures, under the training from scratch protocol, on the challenging UCF101 dataset, surprisingly, with only around \emph{one} percentage of parameters of its manual-design counterparts.
\end{abstract}

\section{Introduction}
\label{sec:intro}
Video action recognition~\cite{le2011learning}, which is a hot topic of video analysis and understanding, has drawn considerable attention from both academia and industry, since it has great value to many potential applications, like behaviour analysis~\cite{keenan2006video}, security, and video affective computing~\cite{peng2019boost}. On one hand, new and large-scale datasets, such as Kinetics~\cite{kay2017kinetics}, Something-Something~\cite{goyal2017something}, make great contribution to video action recognition. On the other hand, video analysis and understanding benefits a lot from the recent advance in deep neural networks, which have already been successfully applied to a large variety of tasks like object detection~\cite{redmon2016you} and machine translation.  However, designing a neural network structure with high performance is a hard work. It requires large amounts of time and efforts of human experts. Specifically, in the action recognition task, to get satisfying performance, the long-range spatial-temporal connections should be taken into consideration. It thus inevitably leads to complicated network architectures and expensive computational costs. Moreover, one has to feed such a network with much training data.

Recently, Neural Architecture Search (NAS)~\cite{zoph2016neural} has shown great superiority over manually designed neural architectures. In image classification, automatically designed architectures like NASNet~\cite{zoph2018learning}, and AmoebaNet~\cite{real2018regularized}, outperform popular human-designed networks, such as VGG~\cite{simonyan2014very}, and ResNet~\cite{he2016deep}. In semantic image segmentation, Auto-Deeplab~\cite{liu2019auto} also performs well even without a pre-train procedure. 
Such promising progresses are mostly benefited from the current strong computational capability, the elaborate search space~\cite{pham2018efficient} as well as the efficiency of search strategies like reinforcement learning~\cite{zoph2016neural}, evolutionary algorithms~\cite{real2018regularized} and gradient methods~\cite{liu2018darts}. However, involved tasks are limited at cases with low-dimensional inputs like text and images.

In this paper, we aim at realizing automatic neural architecture design for action recognition. Fueled by the promising progress of NAS, we propose to introduce NAS into the action recognition task with an efficient fashion. Firstly, a video processing scheme like in temporal segment network (TSN)~\cite{TSN2016ECCV} is adopted to capture global temporal information and reduces the computational cost. It also contributes to the data augmentation. Secondly, to characterize the spatial-temporal dynamics in videos, we provide an efficient solution to search 3D neural operators, by processing the spatial and temporal features separately in a search space with \emph{pseudo 3D}~\cite{tran2018closer} operators. Thirdly, we introduce a directed acyclic graph to model the search space and relax the discrete space into a continuous and a differentiate one~\cite{liu2018darts}. It is thus allowed to achieve highly competitive performance with a small order of magnitude in computational resources against the previous ones with thousands of GPU days to search.


The contributions of this paper are three-fold: Firstly, to our best knowledge, this is the first attempt to automatically design a spatial-temporal neural network for video action recognition issues. Secondly, we design a continuous pseudo 3D neural network search space, where high-efficient operators can be explored with proxyless strategy. Finally, we evaluate the proposed method on UCF101~\cite{Soomro2012UCF101} dataset. Compared with popular action recognition neural architectures without pre-training, our searched architecture gets the best performance with only around one percentage of parameters.

\section{PROPOSED METHODS}
\label{sec:method}
In this section we detail the modularized architecture $\mathcal{C}$ to be searched and the search strategy. Here, we search for a single (2+1)D convolutional module unit and repeat it for multiple times to build a neural network. During the search procedure, we build a shallow networks, $\mathcal{N}(\mathcal{C})$, to explore in the search space.

\begin{figure}[t]
\centering
\includegraphics[width=0.5\textwidth]{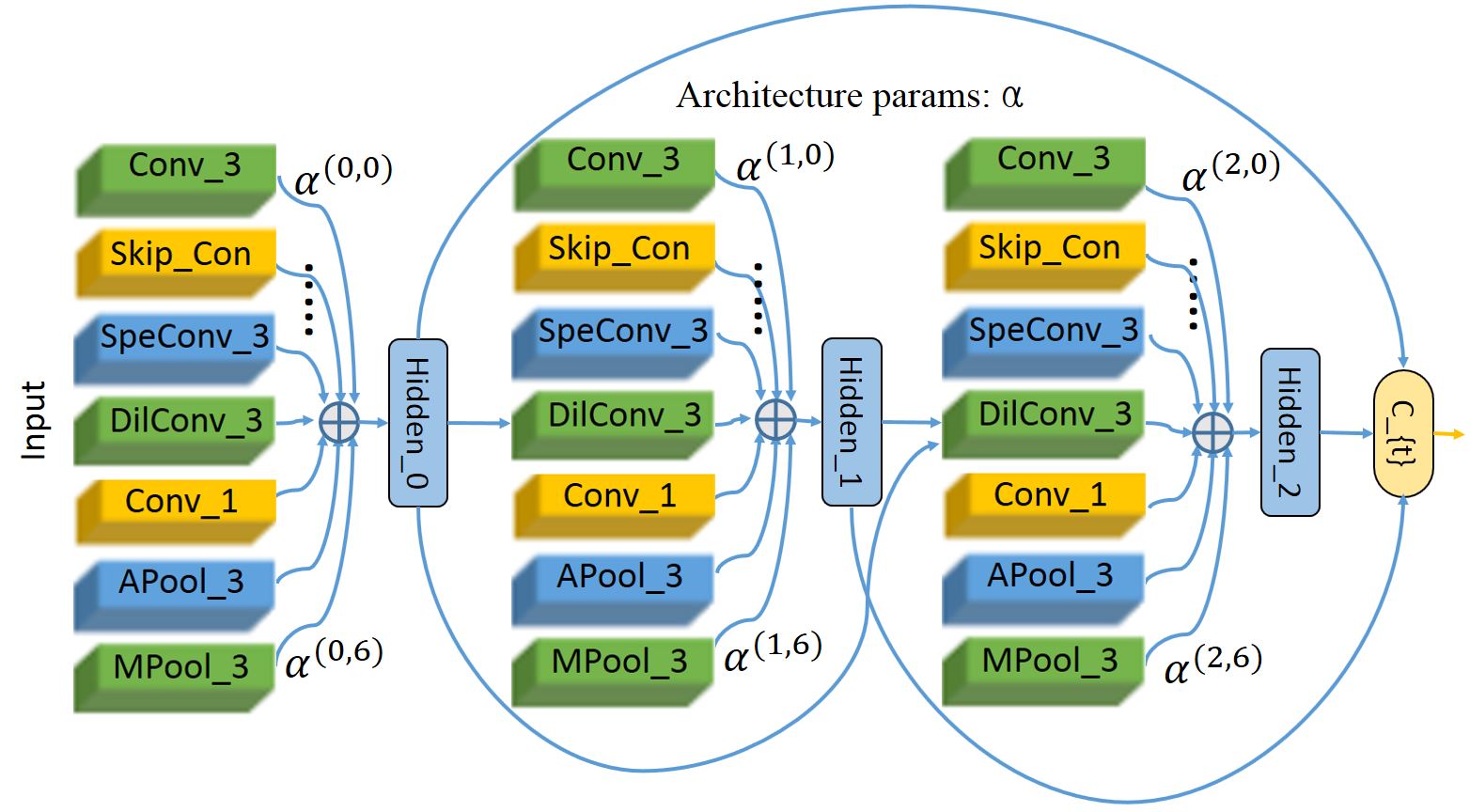}
\caption{\small{Overview of the neural module to be searched. The architecture of the module is determined by different nodes (Hidden\_i) connections and operators. There are three nodes in this module. Concatenating the outputs of them forms the module output. For each connection between nodes, there are eight operator candidates which are detailed at the bottom of Table~\ref{symbol-table}. In a continuous search space, the contribution of each operator is parameterized by a weight $\alpha^{(k,j)}$.  Note that, for interpretability, connections from previous two modules outputs are omitted in this figure.}}
\label{Fig:module}
\vspace{-5mm}
\end{figure}

\begin{table}[t]
\caption{ Symbols and their meaning in the text and figures.} 
\label{symbol-table}

\scalebox{0.9}{\begin{tabular}{ll}
\hline
\hline
Symbol & Meaning \\
\hline
$\mathcal{C}$        & A neural module.\\
$\mathcal{C}\_\{t\}$ & The output of the $t$-th module in a network. \\
Hidden\_i            & The hidden output of the $i$-th node in a module.\\
$\mathcal{N}(\mathcal{C})$        & A neural network build on $\mathcal{C}$.\\
$\Theta$    &  The weights within $\mathcal{N}(\mathcal{C})$.\\
$M$         &  The number of connections in a module.\\
$n$         &  The number of nodes in a module.\\
$\mathcal{O}$ &  A set of operators.\\
$\alpha \in \mathbb{R^{\mathcal{M}\times |\mathcal{O}|}}$    &  The network architecture parameters.\\
$\alpha^{(k,j)}$   &  The $j$-th operator of the $k$-th connection.\\
Conv\_1   &  RELU-CONV-BN block with kernel size 1. \\
Conv\_3   & RELU-CONV-BN block with kernel size 3. \\
SpeConv\_3 &  Separable convolution filters with kernel size $3$.\\
DilConv\_3 & Dilated convolution filters with kernel size $3$.\\
MPool\_3 & Max pooling with kernel size $3$. \\
APool\_3  & Average pooling with kernel size $3$.\\
Skip\_Con  &Skip connection between Nodes.\\
Zero  & No connection between Nodes(Not show in figure).\\
\hline
\end{tabular}}
\vspace{-1em}
\end{table}

 For video action recognition task, a TSN-like scheme is introduced to deal with input clips with various lengths. {It not only reduces computational costs but} also helps {in} capturing global information and data augmentation. We uniformly segment the clip into $N_s$ segments. In every segment, we randomly sample $N_r$ frames, and {regroup} them {with the original order to form} a new clip. {These}  clips of a {fixed length of} ($N_s\times N_r$) are then used as the input to our model.
 
\subsection{Neural module architecture}
As shown in Fig.~\ref{Fig:module}, we build a directed acylic graph to model the neural module unit $\mathcal{C}$. Assume $\mathcal{C}$ of $n$ nodes is the fundamental unit in our neural network for video action recognition. Then the search procedure is to find out the connections between these $n$ nodes and their corresponding transform functions from a set of candidate operators $\mathcal{O}$ to optimize the performance.

\noindent  \textbf{Connections.} Here, connections work as the data flow. For each node, which is the "Hidden\_i" in Fig.~\ref{Fig:module}, we take all the outputs of nodes before it as its inputs. Besides, from the network-level perspective, the outputs from two previous modules are also taken as its inputs (not shown in Fig.~\ref{Fig:module}). {As a result}, there will be totally $M = \frac{(n+1)(n+2)}{2} - 1$ connections {inside a module of $n$ nodes.} For the output of each node, element-wise addition is conducted during its different inputs. Finally, all the outputs of these nodes are concatenated as the module output $\mathcal{C}\_{\{t\}}$.

\noindent \textbf{Operators.} All the connections are filled with one or more operators from $\mathcal{O}$. Here, we decouple every operator into a 2D spatial one and a 1D temporal one since the (2+1)D convolution filters are with less parameters compared to the 3D ones but could get superior performance. In this paper, the initial set of operators in $\mathcal{O}$ {consist of} eight operator candidates, which are listed {in the eight bottom rows} of Table~\ref{symbol-table}. Here, max or average pooling layers and RELU-CONV-BN blocks are prevalent in modern manually designed networks. 'SpeConv\_3' is implemented by three 1D filters. 'DilConv\_3' has a spatial perspective filed of $5\times5$. {For computational consideration, we set all the operators with a relative small kernel size.} 

To find the best operators from $\mathcal{O}$, like \cite{liu2018darts}, we introduce all of them into each connection and model them with the architecture parameters $\alpha$. Then for each connection, element-wise addition is conducted between the outputs of each weighted operator. Therefore, search processing is reduced to find the best architecture parameters $\alpha$ which maximizes the performance on the validation data. At the end of this search, a discrete architecture can be got by functions like softmax.
As a result, the search strategy {is} to adjust the contribution of each operator at each iteration step and find out the network architecture parameter $\alpha = \{\alpha^{(k,j)}\}$, for $k = 0,...,M-1, j = 0,..,|\mathcal{O}|-1~$, which minimizes the task loss{, as formulated as follows:}
\begin{equation}\label{eq:loss}
\alpha^{*} = \arg \min _{\alpha}\L_{valid}(\Theta(\alpha), \alpha). 
\end{equation}
Here, we introduce the parameter sharing scheme in $\mathcal{N}(\mathcal{C})$, thus $\Theta$ is the weights within the network, which is shared by each architecture candidate. $\L_{valid}$ is a loss function {on the validation set}. To better evaluate the architectures, $\Theta$ will also be updated on the training data at each iteration. Let $\L_{train}$ denote the loss function for training. Obviously, the losses are determined by both the module architecture $\alpha$ and network weights $\Theta$. {Solving} 
Eq.~(\ref{eq:loss}) means {to} find the optimal $\alpha$, where the weights $\Theta$ minimize $\L_{train}$. This bilevel problem is hard since they are {interdependent, more precisely, nested.}
We will discuss it in the next section.

\subsection{Search Strategy}
In this subsection, we {elaborate} our searching strategy to optimize {the} neural network architecture. 

{Though the search space is relaxed into a continuous and differentiate one, the optimization of Eq.~(\ref{eq:loss}) is still challenging or even infeasible. The main cause is that} 
for any change of the architecture $\alpha$, {one has to recompute}
$\Theta(\alpha)$ by minimizing $\L_{train}$.
{To alleviate such difficulty, an} alternating iteration approximation {is performed.} 
{More concretely, given a search space,} 
there are two alternative steps to conduct this searching process: Firstly, updating the network weights $\Theta$ by minimizing $\L_{train}$, where the architecture is fixed. Secondly, updating the network architecture $\alpha$ by minimizing $\L_{valid}$, {by fixing the weights of the network.} 
{On this basis, $\Theta$ and $\alpha$ are optimized by alternating between gradient
descent steps in the network weights and the architecture parameters in an iterative manner.}
In fact, this task is very similar with the few-shot meta-learning task~\cite{finn2017model}, which is to adapt a model to a new task using a few data and training process. From this perspective, neural architecture search can be treated as a kind of meta-learning, in which NAS is to transfer the network trained on the training data to adapt the validation data by fine-tuning the architecture. As a result, a surrogate function is employed for $\Theta(\alpha)$ in Eq.~(\ref{eq:loss}). That is

\begin{equation}\label{eq:gradient}
\alpha^{*} = \arg \min_{\alpha}\L_{valid}(\Theta^{'}, \alpha),
\end{equation}
\begin{equation}\label{eq:gradient2}
 \Theta^{'} = \Theta- \epsilon \bigtriangledown_{\Theta}\L_{train}(\Theta, \alpha).
\end{equation}
Here,the $\epsilon$ is a very small value in the search step. By using the gradient descent method with {respect}
to the architecture parameters $\alpha$, we can find a solution of the architecture.
That is

\begin{equation}\label{eq:SGD}
\alpha = \alpha - \gamma \bigtriangledown_{\alpha}.
\end{equation}
Here $\gamma$ is the learning rate, and $\bigtriangledown_{\alpha}$ is calculated by Eq.~{(\ref{eq:2-order})} :

\begin{equation}\label{eq:2-order}
\begin{split}
\bigtriangledown_{\alpha} = &\bigtriangledown_{\alpha}\L_{valid}(\Theta^{'}, \alpha) - \\ 
& \epsilon\bigtriangledown_{\alpha,\Theta}^{2}\L_{train}(\Theta,\alpha)\bigtriangledown_{\Theta^{'}}\L_{valid}(\Theta^{'},\alpha).
\end{split}
\end{equation}
Eqs.~(\ref{eq:SGD}) and~(\ref{eq:2-order}) give us a solution to optimize the architecture by SGD. However, it is not computational efficient since there is a second-order derivative, in which the matrix-vector products are computational expensive. Fortunately, 
Hessian-vector products can be used here to approximate the second-order derivative{~\cite{liu2018darts}, so that the computational complexity can be significantly reduced.}
Besides, it is worth to mention that when $\epsilon \rightarrow{0}$, Eq.~{(\ref{eq:gradient2})} {indicates} that $\Theta^{'} \rightarrow \Theta$, which means a second-order derivative {degenerates} to a first-order {one}. {Therefore}, as a special case of Eq.~{(\ref{eq:2-order})}, one can drop out the one-step unrolled weights operation in Eq.~{(\ref{eq:gradient2})} for computational consideration, which will reduce about one-third of the computational cost, {in case that a certain loss in accuracy is tolerant.} 

\begin{figure*}[thbp!]
\centering
\includegraphics[width=0.6\textwidth]{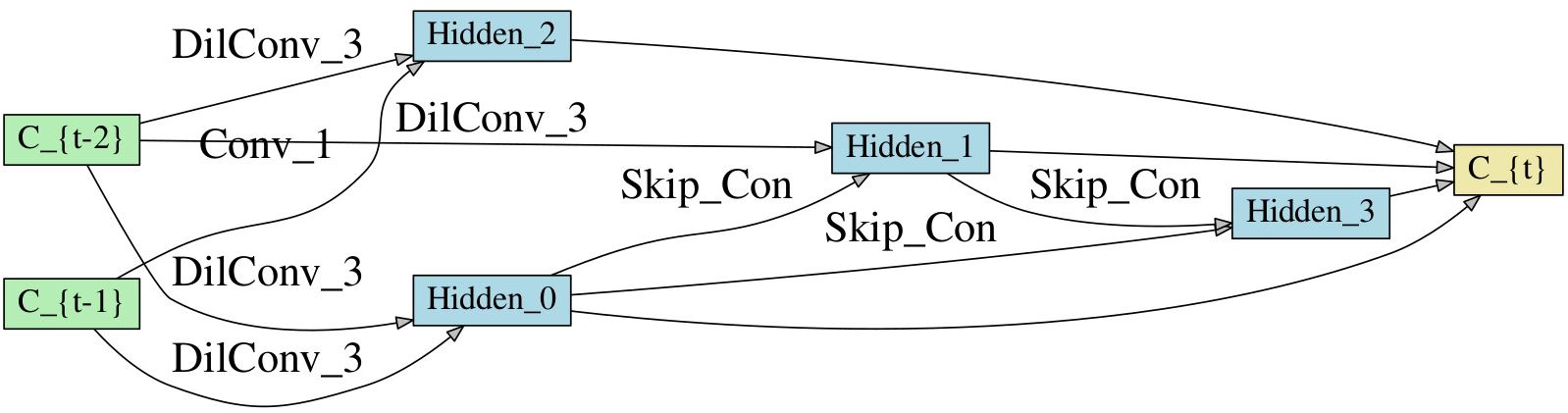}
\caption {\small{Neural architecture searched  by our method. Here, we keep two connections for each node according to $\alpha$. The outputs of two previous modules $\mathcal{C}\_{\{t-1\}}$  and $\mathcal{C}\_{\{t-2\}}$ are taken as inputs for current module. All the outputs of these four nodes are concatenated to form the final output $\mathcal{C}\_{\{t\}}$.}}
\label{Fig:ACT}
\vspace{-5mm}
\end{figure*}

\section{EXPERIMENTS}
In this section, we study the proposed automatic method of designing action recognition network to demonstrate
its advantages over other famous action recognition architectures, e.g., 3D-ResNet \cite{hara2018can}, C3D network \cite{tran2015learning}, and STC-ResNet \cite{diba2018spatio}. We evaluate our algorithm on the challenging
action recognition dataset UCF101, which is a trimmed dataset containing 13320 video clips of 101 classes, with the training from scratch protocol. The training from scratch protocol is commonly used and of great value for issues of limited data or of limited computational resources. Other architectures, such as I3D~\cite{Carreira_2017_CVPR} and TSN, which are either two-stream methods with extra input modality or pre-trained on another larger datasets, are not taken into consideration, owing to the limitation of computational resources.


\textbf{Implementation.} There are two stages to design an automatic action recognition network: training network and updating architecture, which are alternatively performed. We search the network on the \emph{split 1} of UCF101. During the searching processing, we build the network $\mathcal{N}(\mathcal{C})$ with three layers and each layer is constructed by an identical module $\mathcal{C}$, which is the one to be searched. We initialize the first module with four channels and set the number of hidden nodes $n$ to four as a trade-off between accuracy and efficiency. 
For feature inputs with different resolutions, we add an extra feature map adaption layer before the module. Once there need a feature map resolution reduction, the output channels are doubled correspondingly. The network takes eight RGB frames with size $112\times 112$ as inputs. So for the video input pre-processing, we set the segment number $N_{s} = 4$ and number of random sample $N_{r} = 2$. Other data augmentation methods like normalization, random horizontal flip, and random crop are employed. For the objective function, a cross-entropy loss is adopted for the classification task on UCF101. 
For searching the architecture, the whole search procedure is 50 epochs, thus we choose a relative big learning rate. We also decrease the learning rate at each iteration. What makes this search procedure more efficient is that the search model is not thrown away after each validation.

Once we finish the search procedure, we will build and train a final network from scratch. As aforementioned, we construct the network with the returned search result $\alpha$, as illustrated in Fig.~\ref{Fig:module}. Here, for each node, we keep the top-2 connections according to $\alpha$, and then choose the most important operator for each connection based on their contributions. Here, a softmax function is adopted on $\alpha$ to evaluate the contribution of each operator. Then a network can be built by the discrete architecture, which is shown in Fig.~\ref{Fig:ACT}. For the depth of the network, we {investigate} different repeated times, {and find that the network performs best when the depth is set to six. Every two layers, we reduce the feature map resolution by a factor of two along width and height side. Accordingly, we double the channels of feature maps. The whole training process is set as  follows:} 
 We initialize the first module with eight channels and the training iteration is 600 epochs, the learning rate is 0.025, and deceased by a cosine function w.r.t iteration steps. The network is trained with a momentum of 0.9, a weight decay 0.0003, and a mini-batch of 72. For the testing processing, our model predicts a score for each video clip without additional aggregation. For each input clip, we sample eight frames, apply only center crop on the inputs.

\textbf{Result.} 
When the 50 search iteration finished, we choose the best architecture according to their performance. 
We perform the search procedure on a single Nvidia V100 GPU. It takes about 25 hours on the UCF101 to finish this searching process.  Searched architecture is shown in Fig.~\ref{Fig:ACT}. It indicates that the the architecture prefers to the operator with bigger perspective field.

After getting the searched result, we build a network with six layers by repeating this neural module unit. Then, we train on the training data for 600 epochs and test. Table \ref{rerult-table} shows the result of action recognition on UCF101 compared with 3D-convNet, 3D-ResNet 18, 3D-ResNet 101, and different kinds of STC-ResNet networks. 
The result shows that our approach achieves a better accuracy with much fewer parameters than any other models in this table. For instance, 3D-ConvNet, which is a very commonly used architecture, is one hundred times bigger than our network in terms of parameter size. Nevertheless, our model been searched outperforms it by 7\% in terms of recognition accuracy.


\begin{table}[t]
\caption{
Comparison with manually designed 3D networks training from scratch on UCF101 \emph{split 1}. 
As the benchmark codes are not available, we can not get the precise model size of STC-ResNet, though it is clearly larger than ResNet with the same blocks.} 
\label{rerult-table}
\vskip 0.15in
\begin{tabular}{llccr}
\hline
\hline
Architectures & \#params & model size & Accuracy \\
\hline
3D-ResNet  18~\cite{hara2018can}     &33.2M~& 252M & 42.4\% \\
3D-ResNet  101~\cite{hara2018can}    &100M +   & 652M & 46.7\% \\
3D-ConvNet~\cite{tran2015learning}       & 79M & 305M & 51.6\%\\
STC-ResNet  18~\cite{diba2018spatio}    & 33.2M + & - & 42.8\% \\
STC-ResNet  50~\cite{diba2018spatio}    & 92M +   & - &  46.2\%       \\
STC-ResNet 101~\cite{diba2018spatio}   & 100M + & - & 47.9\%\\
\textbf{Ours}    & \textbf{0.67M} & \textbf{7.32M} & \textbf{58.6}\% \\

\hline
\end{tabular}
\vskip -0.1in
\end{table}

\section{Conclusion}
 In this paper, we perform neural architecture search for the 
 action recognition task for the first time. Specifically, we model the neural network by a directed acyclic graph and efficiently search a spatial-temporal neural architecture 
 in a continuous search space.
 We demonstrate that our method outperforms other popular models under the training from scratch protocol, with a surprisingly smaller model size. More concretely, our method improves the accuracy with an over 10\% increase on the UCF101 dataset with approximately \emph{one percentage} of the size of
 famous models like 3D-convNet and 3D-ResNet. In future work, we plan to apply the proposed method to other computer vision tasks.

\section{Acknowledgements}
This work was supported by the Academy of Finland ICT 2023 project (Grant No. 313600), Tekes Fidipro program (Grant No. 1849/31/2015) and Business Finland project (Grant No. 3116/31/2017), Infotech Oulu, and the National Natural Science Foundation of China (Grants No. 61772419). As well, the authors wish to acknowledge CSC-IT Center for Science, Finland, for computational resources.

\bibliographystyle{ieee}
\bibliography{main}

\end{document}

%% file: macros.tex
\newcommand\XP[1]{\textcolor{red}{#1}}
\newcommand\XPC[1]{\textcolor{red}{[Xiaopeng: #1]}}
\newcommand\pw[1]{\textcolor{blue}{#1}}

\newcommand\Zhao[1]{\textcolor{black}{#1}}

\newcommand\up[1]{\textcolor{gray}{#1}}

\def\eg{\emph{e.g.}} \def\Eg{\emph{E.g.}}
\def\ie{\emph{i.e.}}
\def\Ie{\emph{I.e.}}
\def\cf{\emph{c.f.}} \def\Cf{\emph{C.f.}}
\def\etc{\emph{etc.}} \def\vs{\emph{vs.}}
\def\wrt{w.r.t.} 
\def\dof{d.o.f.}
\def\etal{\emph{et al.}}
\def\aka{\emph{a.k.a.}}